\documentclass[journal]{IEEEtran} 

\usepackage{cite}
\usepackage{amsmath,amssymb,amsfonts}
\usepackage{algorithmic}
\usepackage{graphicx}
\usepackage{textcomp}
\usepackage{xcolor}

\usepackage{comment}
\usepackage{mathtools}
\usepackage{gensymb}
\usepackage{mathptmx}
\usepackage{arydshln}
\usepackage{booktabs}
\usepackage{minipage-marginpar}
\usepackage{amsmath,amssymb,exscale}
\usepackage{csquotes}
\usepackage{rotating}
\usepackage{multirow}
\usepackage{array}
\usepackage{subfigure}
\usepackage{adjustbox}
\usepackage{longtable}
\usepackage{scrextend}
\usepackage{graphicx}
\usepackage{wrapfig}
\newcommand{\ie}{\textit{i.e.}}
\newcommand{\eg}{\textit{e.g.}}

\begin{document}

\title{\LARGE \bf
	Place recognition survey: An update on deep learning approaches
}

\author{Tiago Barros, Ricardo Pereira, Lu\'{i}s Garrote, Cristiano Premebida, Urbano J. Nunes

\thanks{The authors are with the University of Coimbra, Institute of Systems and Robotics, Department of Electrical and Computer Engineering, Portugal.
		E-mail:{\tt\small\{tiagobarros,~ricardo.pereira,~garrote,
			~cpremebida,~urbano\}@isr.uc.pt}}
		}
		
\maketitle

\begin{abstract}
Autonomous Vehicles (AV) are becoming more capable of navigating in complex environments with dynamic and changing conditions. A key component that enables these intelligent vehicles to overcome such conditions and become more autonomous is the sophistication of the perception and localization systems. As part of the localization system, place recognition has benefited from recent developments in other perception tasks such as place categorization or object recognition, namely with the emergence of deep learning (DL) frameworks. This paper surveys recent approaches and methods used in place recognition, particularly those based on deep learning. 
The contributions of this work are twofold: surveying recent sensors such as 3D LiDARs and RADARs, applied in place recognition; and categorizing the various DL-based place recognition works into supervised, unsupervised, semi-supervised, parallel, and hierarchical categories. First, this survey introduces key place recognition concepts to contextualize the reader. Then, sensor characteristics are addressed. This survey proceeds by elaborating on the various DL-based works, presenting summaries for each framework. Some lessons learned from this survey include: the importance of NetVLAD for supervised end-to-end learning; the advantages of unsupervised approaches in place recognition, namely for cross-domain applications; or the increasing tendency of recent works to seek, not only for higher performance but also for higher efficiency.
\end{abstract}



\IEEEpeerreviewmaketitle


\section{Introduction}

 Self-driving vehicles are increasingly able to deal with unstructured and dynamic environments, which is mainly due to the development of more robust long-term localization and perception systems. A critical aspect of long-term localization is to guarantee coherent mapping and bounded error over time, which is achieved by finding loops in revisited areas.  Revisited places are detected in long-term localization systems by resorting to approaches such as place recognition and loop closure. Namely, place recognition is a perception based approach that recognizes previously visited places based on visual, structural, or semantic cues. 

\begin{figure}[t]
		\begin{centering}
		\includegraphics[width=1\columnwidth, trim={0.0cm 0cm 0cm 0cm},clip]{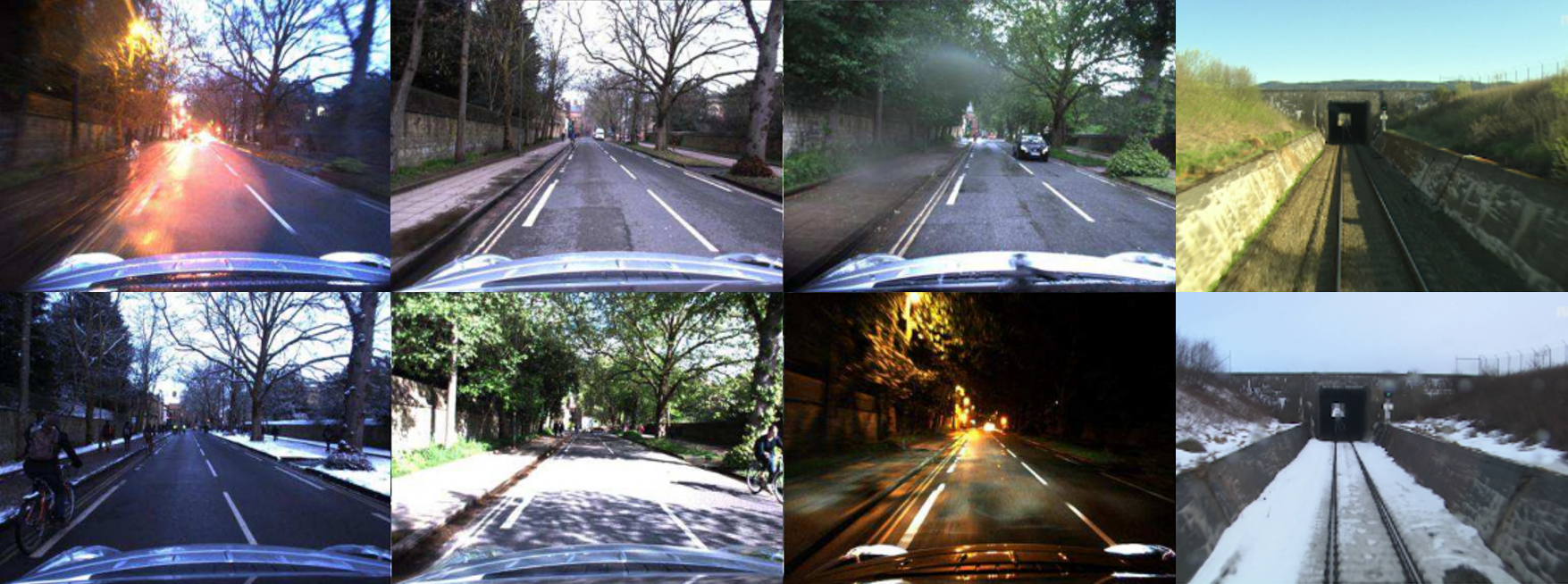}
		\par\end{centering}
	\caption{Illustration of the seasonal environment changes. Images taken from the Oxford Robotcar \cite{RobotCarDatasetIJRR} and Nordland dataset \cite{olid2018single}.}
	\label{fig:seasonal_images}
\end{figure}

Place recognition has been the focus of much research over the last decade. The efforts of the intelligent vehicle and machine vision communities, including those devoted to place recognition, resulted in great achievements, namely evolving towards systems that achieve promising performances in appearance changing and extreme viewpoint variation conditions. Despite the recent achievements, the fundamental challenges remain unsolved, which occur when: 

\begin{itemize}
    \item[$-$] two distinct places look similar (also known as perceptual aliasing);
    \item[$-$] the same places exhibit significant appearance changes over time due to day-night variation, weather, seasonal or structural changes (as shown in Fig. \ref{fig:seasonal_images});
    \item[$-$] same places are perceived from different viewpoints or positions.
\end{itemize}

\noindent Solving these challenges is essential to enable robust place recognition and consequently long-term localization. 

\begin{figure}[t]
		\begin{centering}
		\includegraphics[width=1\columnwidth, trim={0.0cm 0cm 0cm 0cm},clip]{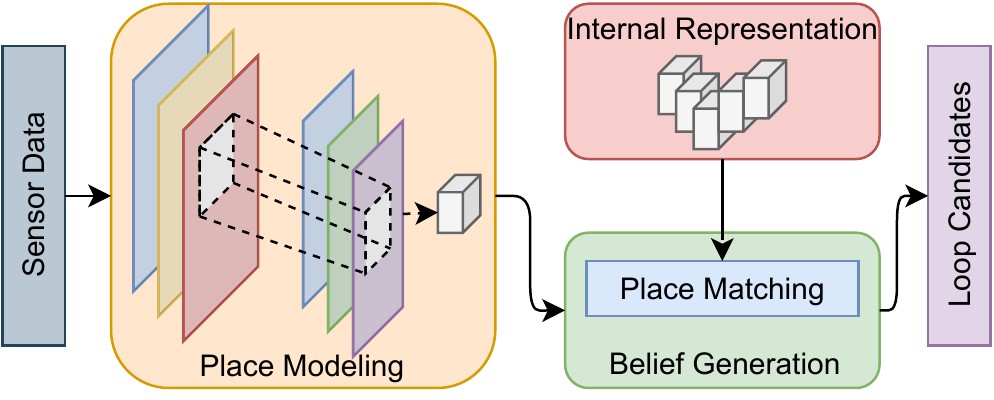}
		\par\end{centering}
	\caption{Generic place recognition pipeline with the following modules: place modeling, belief generation and place mapping. Place modeling creates an internal place representation. Place mapping is concerned with maintaining a coherent representation of places over time. And Belief generation, finally, generates, based on the current place model and the map, loop candidates.}
	\label{fig:pl_block}
\end{figure}

The primary motivation for writing this survey paper is to provide an updated review of the recent place recognition approaches and methods since the publication of previous surveys \cite{lowry2015visual,garcia2015vision}. The goal is, in particular, to focus on the works that are based on deep-learning.

Lowry et al.\cite{lowry2015visual} presented a comprehensive overview of the existing visual place recognition methods up to 2016. The work summarizes and discusses several fundamentals to deal with appearance changing environments and viewpoint variations. However, the rapid developments in deep learning (DL) and new sensor modalities (\eg, 3D LiDARs and RADARs) are setting unprecedented performances, shifting the place recognition state of the art from traditional (handcrafted-only) feature extraction towards data-driven methods.

A key advantage of these data-driven approaches is the end-to-end training, which enables to learn a task directly from the sensory data without requiring domain knowledge for feature extraction.  Instead, features are learning during training, using Convolutional Neural Networks (CNNs). These feature extraction approaches have been ultimately the driving force that has inspired recent works to use supervised, unsupervised, or both learning approaches combined (semi-supervised) to improve performance. The influence of DL frameworks in place recognition is, in particular, observable when considering the vast amount of place recognition works published in the last few years that resort to such methods. 

On the other hand, a disadvantage of DL methods is the requirement of a vast amount of training data. This requirement is in particular critical since the creation of suitable datasets is a demanding and expensive process. In this regard, place recognition has benefited considerably from the availability of autonomous vehicle datasets, which are becoming more and more realistic. Besides more realistic real-world conditions, also data from new sensor modalities are becoming available, for example, new camera types, 3D LiDARs, and, more recently, RADARs \cite{9197298,barnes2020oxford}. This work does not address datasets since this topic is already overviewed in other works such as in \cite{warburg2020mapillary} what place recognition concerns, and in \cite{8667012} broader autonomous driving datasets.

\begin{figure*}[t]
	\begin{centering}
		\includegraphics[width=1\textwidth, trim={0.0cm 0cm 0cm 0cm},clip]{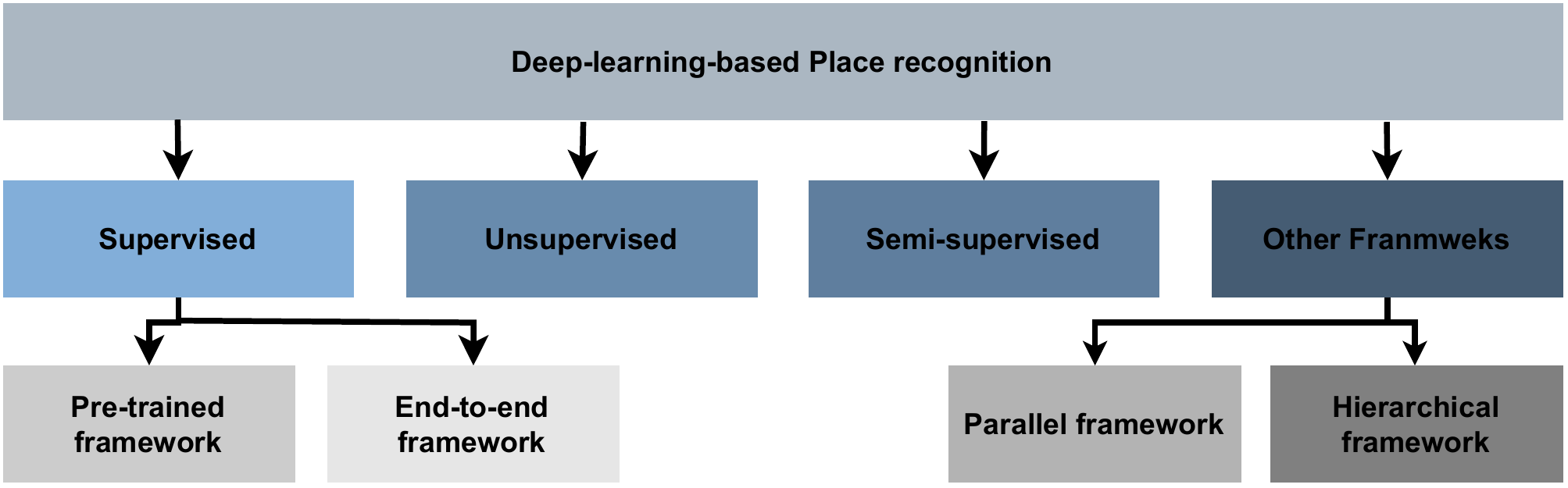}
		\par\end{centering}
	\caption{A taxonomy of recent DL-based place recognition approaches.}
	\label{fig:taxonomy}
\end{figure*}

\begin{figure}[t]
	\begin{centering}
		\includegraphics[width=1\linewidth, trim={0.0cm 0cm 0cm 0cm},clip]{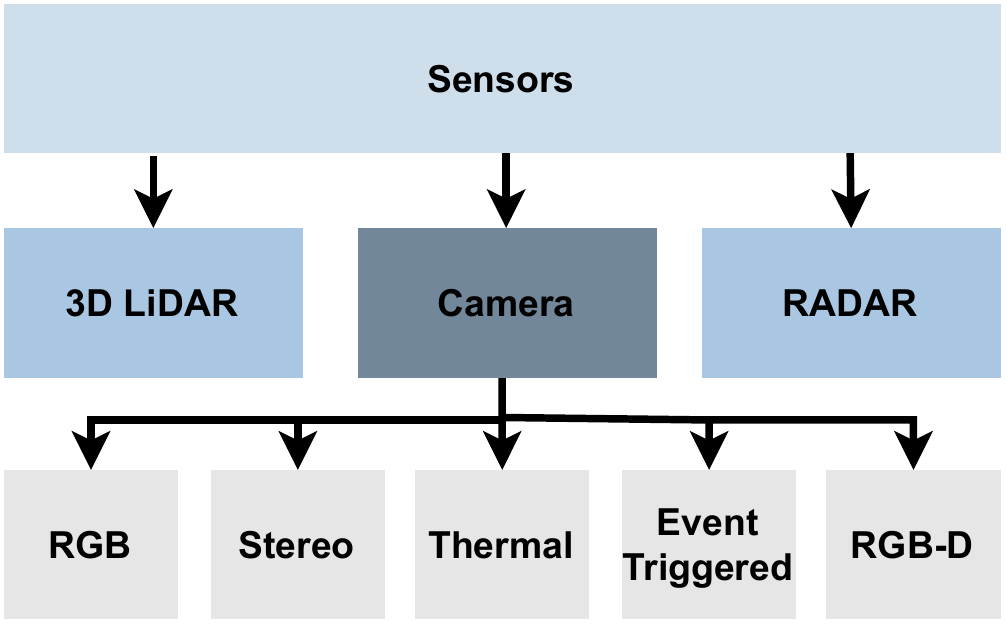}
		\par\end{centering}
	\caption{Popular sensors in place recognition.}
	\label{fig:sensors}
\end{figure}

The contribution of this work is to provide a comprehensive review of the recent methods and approaches, focusing in particular on:
\begin{itemize}
	\item the recent introduced sensors in the context of place recognition, an outline of advantages and disadvantages is presented in Table \ref{tab:sensors} and outline is illustrated in Fig. \ref{fig:sensors};
	\item the categorization of the various DL-based works into supervised, unsupervised, semi-supervised and other frameworks (as illustrated in Fig. \ref{fig:taxonomy}), in order to provide to the reader a more comprehensive and meaningful understanding of this topic. 
\end{itemize}

The remainder of this paper is organized as follows. Section \ref{sec:keyconcepts} is dedicated to the key concepts regarding place recognition. Section \ref{sec:sensors} addresses the supervised place recognition approaches, which include pre-trained and end-to-end frameworks. Section \ref{sec:unsupervised} addresses the unsupervised place recognition approaches. Section \ref{sec:semisupervised} addresses approaches that combine both supervised and unsupervised. Section \ref{sec:other} addresses alternative frameworks that resort to parallel and hierarchical architectures. Lastly, Section \ref{sec:conclusion} concludes the paper.

\section{Key concepts of place recognition}
\label{sec:keyconcepts}

This section introduces the fundamentals and key concepts of place recognition. Most of the concepts here discussed have been already presented in \cite{lowry2015visual}\cite{kostavelis2015semantic}\cite{cadena2016past}\cite{bresson2017simultaneous}, but they are concisely revisited in this section to contextualize the reader and thus facilitate the reading process. 

Thus, before diving into more details, some fundamental questions have to be addressed beforehand. What is a `place' in the place recognition context? How are places recognized and remembered? Moreover, what are the difficulties/challenges when places change over time?

\subsection{What is a place?}

Places are segments of the physical world that can have any given scale - at the limit, a place may represent a single location to an entire region of discrete locations \cite{lowry2015visual} (see examples in Fig. \ref{fig:seasonal_images}). The segments' physical bounds can be defined, resorting to different segmentation criteria: time step, traveled distance, or appearance. In particular, the appearance criterion is widely used in place recognition.  In such a case, a new place is created whenever the appearance of the current location differs significantly from locations that were previously observed \cite{lowry2015visual}. 

\subsection{How are places recognized and remembered?}

Place recognition is the process of recognizing places within a global map, utilizing cues from surrounding environments. This process is typically divided into three modules (as illustrated in Fig.\ref{fig:pl_block}: place modeling, belief generation, and place mapping.

\subsubsection{Place Modeling}

Place modeling is the module that maps the data from a sensor space into a descriptor space. Sensory data from cameras, 3D LiDARs \cite{8593953,8500682} or RADARs \cite{gadd2020look} are used to model the surrounding environment, which is achieved by extracting meaningful features.
 
Feature extraction approaches have evolved immensely over the last decade. Classical approaches rely on handcrafted descriptors such as SWIFT\cite{790410}, SURF\cite{bay2006surf}, Multiscale Superpixel Grids \cite{7381637}, HOG\cite{dalal2005histograms}  or  bag-of-words\cite{1467486}, which are mainly build based on the knowledge of domain experts (see \cite{lowry2015visual} for further understanding). On the other hand, DL-based techniques, namely CNNs, are optimized to learn the best features for a given task \cite{lecun2015deep}. With the increasing dominance of DL in the various perception tasks, also place recognition slowly benefited from these techniques, using initially pre-trained models from other tasks (\eg, object recognition \cite{yue2015exploiting} or place categorization \cite{sunderhauf2015performance,garg2018don}), and more recently using end-to-end learning techniques trained directly on place recognition tasks \cite{arandjelovic2016netvlad,9140362}.  
 
\subsubsection{Place Mapping}

Place mapping refers to the process of maintaining a faithful representation of the physical world. To this end, place recognition approaches rely on various mapping frameworks and map update mechanisms. Regarding the mapping frameworks, three main approaches are highlighted: database \cite{schonberger2018semantic,garg2019look}, topological \cite{paul2010fab,korrapati2012image,9010336} or topological-metric \cite{dayoub2011long,churchill2013experience}.

Database frameworks are abstract map structures, which store arbitrary amounts of data without any relation between them. These frameworks are mainly used in pure retrieval tasks and resort, for retrieval efficiency, to   k-dimensional \cite{1307183,8968094}, Chow Liu trees \cite{cummins2008fab} or Hierarchical Navigable Small
World (NSW) \cite{malkov2018efficient} to accelerate nearest neighbor search.

Topological(-metric) maps, on the other hand,  are graph-based frameworks, which represent the map through nodes and edges. The nodes represent places in the physical world, while the edges represent the relationships among the nodes (\eg, the similarity between two nodes). A node may represent one, or several locations, defining in the latter case a region in the physical world.  The topological-metric map differs from pure topological in respect of how nodes relate \ie, while in pure topological maps no metric information is used in the edges; in topological-metric maps, nodes may relate with other nodes through relative position, orientation, or metric distance \cite{lowry2015visual}. An example of such a mapping approach is the  HTMap approach \cite{7938750}.

Regarding map updating, database frameworks usually are not updated during operation time, while topological frameworks can be updated. Update strategies include simple methods, which update nodes as loops occur\cite{paul2010fab}, or more sophisticated ones, where long- short-term memory-based methods are used \cite{6907255}. 

\subsubsection{Belief Generation}

The belief generation module refers to the process of generating a belief distribution, which represents the likelihood or confidence of the input data matching a place in the map. This module is thus responsible to generate loop candidates based on the belief scores, which can be computed using methods based on frame-to-frame \cite{8972582,8202131,garg2018lost}, sequence of frames\cite{milford2012seqslam,9128035,zhang2016robust,gadd2020look}, hierarchical, graphs\cite{9196906} or probabilistics\cite{8618373,6943207,6177276}. 

The frame-to-frame matching approach is the most common in place recognition. This approach usually computes the belief distribution by matching only one frame at the time; and uses KD trees \cite{1307183,8968094} or Chow Liu trees \cite{cummins2008fab} for nearest neighbor search, and cosine \cite{garg2018lost}, Euclidean distance \cite{liu2019lpd}, Hamming distance \cite{8968599} to compute the similarity score.

On the other hand, sequence-based approaches compute the scores based on sequences of consecutive frames, using, for example, cost flow minimization \cite{7989671}  to find matches in the similarity matrix. Sequence matching is also implementable in a probabilistic framework using Hidden Markov Models \cite{6943207} or Conditional Random Fields \cite{6177276}. 

Hierarchical methods combine multiple matching approaches in a single place recognition framework. For example, the coarse-to-fine architecture \cite{liu2019seqlpd,garg2019look} selects top candidates in a coarse tier, and from those, selects the best match in a fine tier.

\subsection{What are the major challenges?}

Place recognition approaches are becoming more and more sophisticated as the environment and operation conditions become more similar to real-world situations. An example of this is the current state-of-the-art of place recognition approaches, which can operate over extended areas in real-world conditions with unprecedented performances. Despite these achievements, the major place recognition challenges remain unsolved, namely places with similar appearances; places that change in appearance over time; places that are perceived from different viewpoints; and scalability of the proposed approaches in large environments.

\subsubsection{Appearance Change and Perceptual Aliasing}

Appearance-changing environments and perceptual aliasing have been in particular the focus of much research.  As autonomous vehicles operate over extended periods, their perception systems have to deal with environments that change over time due to for example different weather or seasonal conditions or due to structural changes. While the appearance changing problem is originated when the same place changes over time in appearance, perceptual aliasing is caused when different places have a similar appearance. These conditions affect in particular place recognition since the loop decisions are affected directly by the appearance.  

A variety of works have been addressing these challenges from various perspectives. From the belief generation perspective, sequence-based matching approaches \cite{8633431,7368110,milford2012seqslam,7759667,6177276} are highlighted as very effective in these conditions. Sequence matching is the task of aligning a pair of a template and query sequences, which can be implemented through minimum cost flow \cite{milford2012seqslam,Naseer2014robust}, or probabilistically using Hidden Markov models\cite{ 6943207} or Conditional Random Fields\cite{6177276}.  Another way is to address this problem from the place modeling perspective: extracting condition-invariant features \cite{8793752,7759095}, extracting for example features from the middle layers of CNN's \cite{sunderhauf2015performance}. On the other hand, matching quality of descriptors can be improved through descriptors normalization\cite{garg2018don,naseer2017semantics} or through  unsupervised techniques such as dimensionality reduction \cite{6386145}, change removal \cite{7468579}, K-STD  \cite{9197044}  or delta descriptors \cite{9128035}.

\subsubsection{Viewpoint Changing }

Revisiting a place from different viewpoints - at the limit opposite direction (180º viewpoint variation)\cite{garg2018don} - is also challenging for place recognition. That is, in particular, true for approaches that rely on sensors with a restricted field-of-view (FoV) or without geometrical sensing capabilities. When visiting a place, these sensors only capture a fraction of the environment, and when revisiting from a different angle or position, the appearance of the scene may differ or even additional elements may be sensed, generating a complete different place model.  

To overcome these shortcomings, visual-based approaches have resorted to semantic-based features\cite{garg2018lost,garg2019semantic}. For example extracting features from higher-order CNN layers, which have a semantic meaning, have demonstrated to be more robust to viewpoint variation \cite{sunderhauf2015performance}. Other works propose the use of panoramic cameras \cite{arroyo2014bidirectional} or 3DLiDAR \cite{angelina2018pointnetvlad}, being thus irrelevant what orientation places are perceived in future visits. Thus relying on sensors and methods that do not depend on orientation (also called viewpoint-invariant) turn place recognition more robust.  

\subsubsection{Scalability} 

Another critical factor of place recognition is concerned with scalability \cite{8253815,9009095,8968599,do2019compact,9196522,9009095,9196827}. As self-driving vehicles operate in increasingly larger areas, more places are visited and maps become larger and larger, increasing thus computational demand, which affects negatively the inference efficiency. Thus, to boost inference efficiency,  approaches include: efficient indexing \cite{cummins2011appearance,7811208}, hierarchical searching \cite{mohan2015environment,mactavish2014towards}, hashing \cite{8968599,do2019compact,8019479,sunderhauf2015performance,9196827}, scalar quantization \cite{9196827},  Hidden Markov Models (HMMs) \cite{9009095,9196522} or learning regularly repeating visual patterns \cite{8253815}. For example in \cite{9196827} a hashing-based approach is used in a visual place recognition task with a large database to both maintain the storage footprint of the descriptor space small and boost retrieval.

\section{Sensors}
\label{sec:sensors}

An important aspect of any perception-based application is the selection of appropriate sensors.  To this end, the selection criterion has to consider the specificities of both the application and the environment for the task in hand. In place recognition, the must used sensors are cameras \cite{schonberger2018semantic,garg2019look,garg2019semantic,arandjelovic2016netvlad,garg2018lost,garg2018don,milford2012seqslam,hansen2014visual,churchill2013experience}, LiDARs \cite{dube2020segmap,rizzini2019geometric,angelina2018pointnetvlad,paul2010fab,premebida2017dynamic,cao2018robust,8618373,8968094,8500682} and RADARs \cite{gadd2020look,suaftescu2020kidnapped,tang2020rsl,7487687}. Although in a broader AV context, these sensors are widely adopted \cite{8585340,6728344}, in place recognition, cameras are the most popular in the literature, followed by  LiDARs, while RADARs are a very recent technology in this domain. For the remaining of this section, each sensor is detailed and an outline is presented in Table\,\ref{tab:sensors}.

In the place recognition literature, cameras are by far the most used sensor. The vision category includes camera sensors such as monocular \cite{schonberger2018semantic}, stereo \cite{cadena2010robust}, RGBD \cite{morris2014multiple}, thermal \cite{9197514} or event-triggered \cite{fischer2020event}.  Cameras provide dense and rich visual information, which can be provided at a high frame rate (ranging up to 60Hz) with a relatively low cost. On the other hand, vision data is very sensitive when faced with visual appearance change and viewpoint variation, which is a tremendous disadvantage compared with the other modalities. 
Besides vision data, cameras are also able to return depth maps. This is achieved either with RGB-D\cite{yu2017place,morris2014multiple}, stereo cameras\cite{cadena2010robust}, or trough structure from motion (SfM)\cite{scaramuzza2010closing} methods. In outdoor environments, the limited field-of-view (FoV) and noisy depth measurements are a clear disadvantage when compared with the depth measurements of recent 3D LiDARs.   

LiDAR sensors gained more attention, in place recognition, with the emergence of the 3D rotating version. 3D LiDARs capture the spatial (or geometrical) structure of the surrounding environment in a single 360\degree swift, measuring the time-of-flight (ToF) of reflected laser beams. These sensors have a sensing capacity of up to 120m with a frame rate of 10 to 15 Hz. Such features are particularly suitable for outdoor environment, since measuring depth through ToF is not influenced by lighting or visual appearance conditions. This is a major advantage when compared with cameras. On the other hand,  disadvantages are related to the high cost and the large size, which have been promised to be surpassed by the solid-state versions. An additional weak point is the sensitiveness of this technology towards the reflectance property of objects. For example, glass, mirror, smoke, fog, and dust reduce sensing capabilities.

Radar sensors measure distance through time delay or phase shift of radio signals, which makes them very robust to different weather or lighting conditions. The reasonable cost and long-range capability \cite{sless2019self} are features that are popularizing radars in tasks such as environment understanding \cite{dickmann2016automotive} and place recognition \cite{gadd2020look}.  However, radars continue to face weaknesses in terms of low spatial resolution and interoperability \cite{dickmann2016automotive}, disadvantages when compared with LiDARs or cameras.

\begin{table}[tp]
  \centering
  \caption{Sensors for place recognition: pros and cons.}
    \begin{tabular*}{\columnwidth}{p{0.03\columnwidth}p{0.4\columnwidth}p{0.45\columnwidth}}
    \toprule
    \textbf{Sensor} & \multicolumn{1}{c}{\textbf{Advantage}} & \multicolumn{1}{c}{\textbf{Disadvantage}} \\
    \midrule
    \midrule
    \parbox[t]{2mm}{\multirow{5}{*}{\rotatebox[origin=c]{90}{Camera}}} & - Low cost\newline{}- Dense color information\newline{}- Low energy consumption\newline{}- High precision/resolution\newline{}- High frame rate & - Short range\newline{}- Sensitive to light\newline{}- Sensitive to calibration\newline{}- Limited  FoV\newline{}- Difficulty in textureless environment \\
    \midrule
    \parbox[t]{2mm}{\multirow{5}{*}{\rotatebox[origin=c]{90}{3D LiDAR}}}   &  - Long range\newline{}- 360º FoV\newline{}- Robust to appearance changing conditions\newline{}-high precision/resolution  & - High cost\newline{}- Sensitive to reflective and foggy environments\newline{}- Bulky\newline{}- Fragile mechanics \\
    \midrule
    \parbox[t]{2mm}{\multirow{5}{*}{\rotatebox[origin=c]{90}{RADAR}}}   & - Low cost\newline{}- Very Long range\newline{}- Precise velocity estimation\newline{}- Insensitive to weather conditions & - Narrow FoV\newline{}- Low resolution \\
    \bottomrule
    \end{tabular*}%
  \label{tab:sensors}%
\end{table}%


 \begin{figure}[t]
	\centering
	\includegraphics[width=1\linewidth]{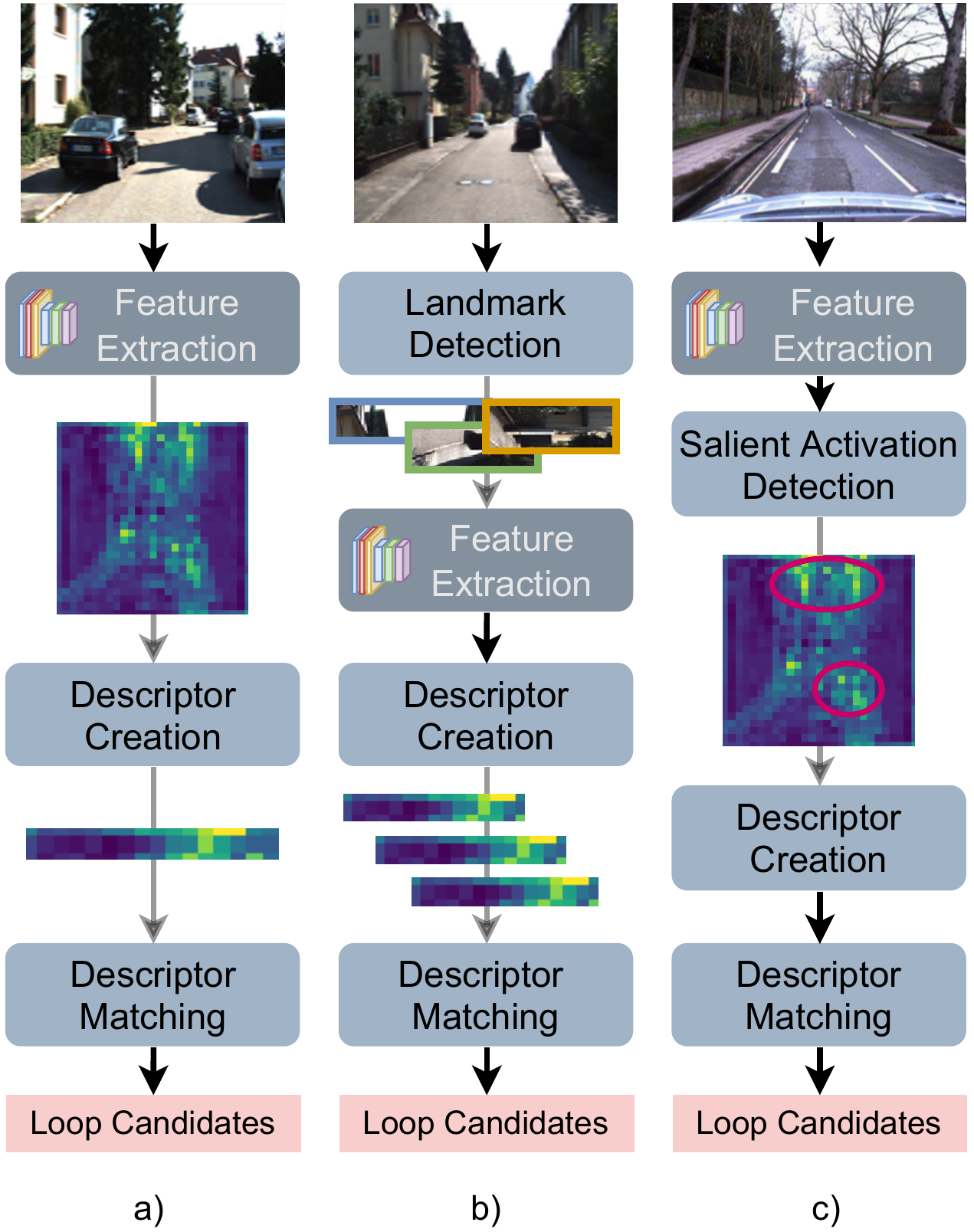}
	\caption{Block diagram of pre-trained frameworks a) holistic-based  b) landmark-based and c) region-based.}
	\label{fig:pretrained}	
\end{figure}

\section{Supervised place recognition} \label{sec:supervised}

This section addresses the place recognition approaches that resort to supervised deep learning. Supervised machine learning techniques learn a function that maps an input representation (\eg, images, point clouds) into an output representation (\eg categories, scores, bounding boxes, descriptor) utilizing labeled data. In deep learning, this function assumes the form of weights in a network with staked layers. The weights are learned progressively by computing the error between predictions and ground-truth in a first step, and in a second step, the error is backpropagated using gradient vectors\cite{lecun2015deep}.  This procedure (i.e., error measuring and weight adjusting) is repeated until the network's predictions achieve adequate performance.  The advantage of such a learning process, particularly when using convolutional networks (CNN), is the capability of automatically learning features from the training data, which, in classical approaches, required a considerable amount of engineering skill and domain expertise. On the other hand, the disadvantages are related to the necessity of a massive amount of labeled data for training, which is expensive to obtain\cite{merrill2018lightweight}.

In place recognition, deep supervised learning enabled breakthroughs. Especially, the capability of CNNs to extract features led to more descriptive place models, improving place matching. Early approaches relied mostly on pre-trained (or off-the-shelf) CNNs that were trained on other vision tasks\cite{yue2015exploiting,sunderhauf2015performance}. But more recently, new approaches enabled the training of DL networks directly on place recognition tasks in a end-to-end fashion \cite{arandjelovic2016netvlad,angelina2018pointnetvlad,DBLP:conf/icra/SaftescuGMBN20,yu2019spatial}.   

\subsection{Pre-trained-based Frameworks}

\begin{table*}[t]
	\centering
	\caption{Summary of recent works on supervised place recognition using pre-trained frameworks. All the works use camera-based data. BG = Belief Generation and PM = Place mapping. }
	{\renewcommand{\arraystretch}{2}
		\begin{adjustbox}{max width=\textwidth}
			\begin{tabular}{p{0.04\textwidth}p{0.04\textwidth}p{0.4\textwidth}p{0.2\textwidth}p{0.3\textwidth}}
				\toprule
				\multicolumn{1}{c}{\textbf{Type}} & \multicolumn{1}{c}{\textbf{Ref}} & \multicolumn{1}{c}{\textbf{Model}} & \multicolumn{1}{c}{\textbf{BG/PM}} & \multicolumn{1}{c}{\textbf{Dataset}} \\
				\midrule
				\midrule
				\multirow{6}{*}{\rotatebox[origin=c]{90}{Holistic-based}} & \cite{yue2015exploiting}	& Feature Extraction:  OxfordNet \cite{Simonyan15} and GoogLeNet\cite{7298594}; \newline{} Descriptor:  VLAD \cite{jegou2011aggregating} + PCA \cite{jegou2012negative} &	 L2 distance/Database &	Holidays \cite{jegou2008hamming}; Oxford \cite{philbin2007object}; \newline{} Paris \cite{philbin2008lost} \\
				& \cite{7759685}  &	Feature Extraction:  CNN-VTL (VGG-F\cite{Chatfield2014devil}) \newline{} Descriptor:  Conv5 layer + Random selection (LDB\cite{yang2013local}) &	Hamming distance/Database &	Nordland \cite{olid2018single}; CMU-CVG Visual Localization \cite{badino2012real}; Alderley \cite{7989671}; \\
				& \cite{sunderhauf2015performance} & Feature Extraction: AlexNet \cite{krizhevsky2017imagenet}; \newline{}Descriptor: Conv3 layer & Hamming KNN/Database & Nordland\cite{olid2018single}; Gardens Point\cite{sunderhauf2015performance}; The Campus Human vs. Robot; The St. Lucia \cite{glover2010fab} \\
				\midrule
				\multirow{6}{*}{\rotatebox[origin=c]{90}{Landmark-based}}  & \cite{garg2018don} & Landmark Detection: Left and right image regions \newline	Feature Extraction: CNN Places365 \cite{zhou2017places} \newline Descriptor:  fc6 + normalization + concatenation &	Sequence Match/Database & Oxford Robotcar \cite{philbin2007object}; \newline  University Campus; \\
				& \cite{sunderhauf2015place} &	Landmark Detection: Edge Boxes \newline Feature Extraction:  ALexNet \newline{} Descriptor: Conv3 layer + Gaussian Random Projection \cite{Sanjoy2000radom} &	Cosine KNN/Database	 & Gardens Point \cite{sunderhauf2015performance}; Mapillary; \newline  Library Robot Indoor; Nordland \cite{olid2018single}; \\
				& \cite{8653820} & Landmrk detection:  BING \cite{6909816} \newline Feature Extraction: AlexNet \newline Descriptor: pool 5 layer +  Gaussian Random Projection \cite{bingham2001random,Sanjoy2000radom}  + normalization &	L2- KNN/Database  &	 Gardens Point\cite{sunderhauf2015performance}; Berlin A100,  Berlin Halenseestrasse  and Berlin Kudamm \cite{sunderhauf2015place};  Nordland  \cite{olid2018single};  St. Lucia  \cite{glover2010fab}; \\
				\midrule
				\multirow{6}{*}{\rotatebox[origin=c]{90}{Region-based}} & \cite{naseer2017semantics} &	Feature Extraction: Fast-Net (VGG) \cite{7759717} \newline{} Descriptor:  conv3 +  L2-normalization + Sparse Random Projection \cite{achlioptas2003database} &	Cosine distance/Database &	Cityscapes \cite{7780719}; \newline{} Virtual KITTI  \cite{7780839}; \newline{} Freiburg; \\
				& \cite{8202131} & Feature Extraction: VGG16\cite{simonyan2014very}  \newline{} Descriptor:  Salient regions from different layers + Bag-of-Words \cite{sivic2003video} & Cross matching/ Database & Gardens Point \cite{sunderhauf2015performance}; Nordland \cite{olid2018single};  Berlin A100,  Berlin Halenseestrasse  and Berlin Kudamm \cite{sunderhauf2015place};  \\
				&	\cite{khaliq2019holistic} &  Feature Extraction: AlexNet365\cite{chen2017only} \newline  Descriptor:  (Region-VLAD)  salient regions + VLAD & Cosine distance/Database	& Mapillary;  Gardens Point \cite{sunderhauf2015performance}; Nordland \cite{olid2018single}; Berlin A100,  Berlin Halenseestrasse  and Berlin Kudamm \cite{sunderhauf2015place};\\
				\bottomrule
			\end{tabular} 
		\end{adjustbox}
	}
	\label{tab:pretrained}%
\end{table*}%

In this work, pre-trained place recognition frameworks refer to  approaches that extract features from pre-trained CNN models, which are originally trained on other perception tasks (\eg, object recognition \cite{yue2015exploiting},  place categorization \cite{sunderhauf2015performance,garg2018don} or segmentation \cite{naseer2017semantics}). Works using such models fall into three categories: holistic-based, landmark-based, and region-based. Figure \ref{fig:pretrained} illustrates such approaches applied to an input image.

\subsubsection{Holistic-based}

Holistic approaches refer to works that feed the whole image to a CNN and use all activations from a layer as a descriptor.  The hierarchical nature of CNNs makes that the various layers contain features with different semantic meanings. Thus, to assess which layers generate the best features for place recognition, works have conducted ablation studies, which compared the performance of the various layers towards appearance and viewpoint robustness and compared the performance of object-centric, place-centric, and hybrid networks (i.e., networks trained respectively for object recognition, place categorization and both).  Moreover, as CNN layers tend to have many activations, the proposed approaches compress the descriptor to a more tractable sized for efficiency reasons.

Ng et al. \cite{yue2015exploiting} study the performance of each layer, using pre-trained object-centric networks such as OxfordNet \cite{Simonyan15} and GoogLeNet\cite{7298594} to extract feature from images. The features are encoded into VLAD descriptors and compressed using PCA \cite{jegou2012negative}. Results show that performance increases as features are extracted from deeper layers, but drops again at the latest layers.  Matching is achieved by computing the L2 distance of two descriptors.

A similar conclusion is reached by Sünderhauf et al. \cite{sunderhauf2015performance}, using holistic image descriptors extracted from AlexNet\cite{krizhevsky2012imagenet}. Authors argue that the semantic information encoded in the middle layers improves place recognition when faced with severe appearance change, while features from higher layers are more robust to viewpoint change. The work further compares AlexNet (object-centric) with  Places205 and Hybrid \cite{zhou2014learning}, both trained on a scene categorization task (i.e., place-centric networks) \cite{zhou2014learning}, concluding that, for place recognition, place-centric networks outperform  object-centric CNNs. The networks are tested using a cosine-based KNN approach for matching, but for efficiency reason, the cosine similarity was approximated by the Hamming distance \cite{ravichandran2005randomized}.

On the other hand, Arroyo et al. \cite{7759685} fuse features from multiple convolutional layers at several levels and granularities and show that this approach outperforms approaches that only use features from a single layer. The CNN architecture is based on the VGG-F \cite{Chatfield2014devil}, and the output features are further compressed using a random selection approach for efficient matching.

\subsubsection{Landmark-based}

Landmark-based approaches, contrary to the aforementioned methods, do not feed the entire image to the network; instead, these approaches use, in a pre-processing stage, object proposal techniques to identify potential landmarks in the images, which are feed to the CNN. Contrary to holistic-based approaches, where all image features are transformed to descriptors, in landmark-based approaches, only the features from the detected landmarks are converted to descriptors. Detection approaches used in these works include  Edge Boxes, BING, or simple heuristics.

With the aim of addressing the extreme appearance and viewpoint variations problem in place recognition, Sünderhauf et al. \cite{sunderhauf2015place} propose such a landmark detection approach. Landmarks are detected using  Edge Boxes \cite{zitnick2014edge} and are mapped into a feature space using the features from Alexnet's\cite{krizhevsky2012imagenet}  conv3 layer.  The descriptor is also compressed for efficiency reasons, using a Gaussian Random Projection approach \cite{Sanjoy2000radom}.  

A similar approach is proposed by Kong et al. \cite{8653820}. However, instead of detecting landmarks using  Edge Boxes \cite{zitnick2014edge} and extracting features from conv3 layer of Alexnet,  landmarks are detected using   BING \cite{6909816} and features are extracted from a pooling layer. 

A slightly different approach is proposed by Garg et al. \cite{garg2018don}, which resorting to Places365 \cite{zhou2017places}, also highlights the effectiveness of place-centric semantic information in extreme variations such as front versus rear view. In particular, this work crops the right and left regions of images, which has been demonstrated to possess useful information\cite{naseer2017semantics}, for place description. The work highlights the importance of semantics-aware features from higher-order layers for viewpoint and condition invariance. Additionally, to improve robustness against appearance,  a descriptor normalization approach is proposed. Descriptor normalization of the query and reference descriptors are computed independently since the image conditions differ (i.e., day-time vs. night-time).  While matching is computed using SeqSLAM \cite{milford2012seqslam}.

\begin{figure*}[t]
	\centering
	\includegraphics[width=1\linewidth]{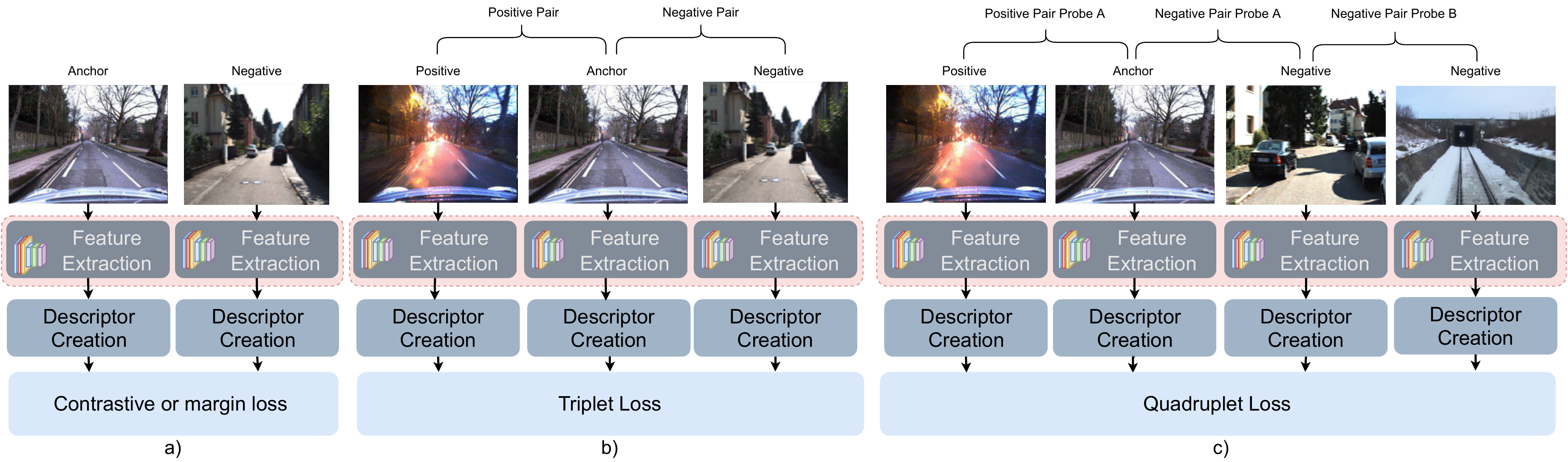}
	\caption{Block diagram of training strategies using a) contrastive-based and margin-based, b) triplet and c) quadruplet loss function.}
	\label{fig:loss}	
\end{figure*}

\subsubsection{Region-based}

Region-based methods, similarly to landmark-based approaches, rely on local features; however, instead of utilizing object proposal methods, the regions of interest are identified on the CNN layers, detecting salient layer activations. Therefore, region-based methods feed the entire image to the DL model and use, for the descriptors, only the salient activation in the CNN layers.

Addressing the problem of viewpoint and appearance changing in place recognition, Chen et al. \cite{8202131} propose such a region-based approach that extracts salient regions without relying on external landmark proposal techniques.  Regions of interest are extracted from various CNN layers of a pre-trained VGG16\cite{simonyan2014very}. The approach extracts explicitly local features from the early layers and semantic information from later layers. The extracted regions are encoded into a descriptor, using a bag-of-words-based approach \cite{sivic2003video} which is matched using a cross-matching approach.  

Naseer et al. \cite{naseer2017semantics}, on the other hand, learn the activation regions of interest resorting to segmentation. In this work, regions of interest represent stable image areas, which are learned using Fast-Net \cite{7759717}. Fast-Net is an up-convolutional Network that provides near real-time image segmentation.   Due to being too large for real-time matching,  the features resulting from the learned segments are encoded into a lower dimensionality using L2-normalization and Sparse Random Projection \cite{achlioptas2003database}. This approach, in particular, learns human-made structure due to being more stable for more extended periods.

With the aim of reducing the memory and computational cost, Khaliq et al. \cite{khaliq2019holistic} propose Region-VLAD. This approach leverages a lightweight place-centric CNN architecture (AlexNet365\cite{chen2017only}) to extract regional features. These features are encoded using a VLAD method, which is specially adapted to gain computation-efficiency and environment invariance.

\begin{table*}[t]
	\centering
	\caption{Summary of recent works on supervised end-to-end place recognition. BG = Belief Generation and PM = Place mapping. }
	{\renewcommand{\arraystretch}{2}
		\begin{adjustbox}{max width=\textwidth}
			\begin{tabular}{p{0.04\textwidth}p{0.04\textwidth}p{0.3\textwidth}p{0.2\textwidth}p{0.13\textwidth}p{0.2\textwidth}}
				\toprule
				\multicolumn{1}{c}{\textbf{Sensor}}  & \multicolumn{1}{c}{\textbf{Ref}} & \multicolumn{1}{c}{\textbf{Architecture}} & \multicolumn{1}{c}{\textbf{Loss Function}} & \multicolumn{1}{c}{\textbf{BG}/\textbf{PM}} & \multicolumn{1}{c}{\textbf{Dataset}}\\
				\midrule
				\midrule
				\multirow{11}{*}{\rotatebox[origin=c]{90}{Camera}} & \cite{arandjelovic2016netvlad}  &\textbf{NetVLAD:}  \newline{} VGG/AlexNet + NetVLAD layer  &  Triplet loss &  KNN /Database & Google Street View Time Machine; Pitts250k \cite{7054472}; \\
				&  \cite{9140362} & 2D CNN visual and 3D CNN structural feature extraction + Feature fusion network;  &  Margin-based loss \cite{8237571} & KNN /Database  & Oxford RobotCar \cite{RobotCarDatasetIJRR}; 
				\\
				& \cite{yu2019spatial}  &\textbf{ SPE-VLAD:}  \newline{} (VGG-16 network or ResNet18)  + spatial pyramid structure + NetVLAD layer  &  Weighted Triplet & L2 /Database & Pittsburgh \cite{7054472};\newline{}TokyoTimeMachine\cite{Torii-CVPR2015};\newline{}Places365-Standard \cite{zhou2017places}; 
				\\
				& \cite{qiu2018siamese} &\textbf{Siamese-ResNet:} \newline{} ResNet in the siamese network &  L2-based loss \cite{chopra2005learning} & L2 /Database &  TUM \cite{6385773}; 
				\\
				& \cite{8968599} &	 MobileNet \cite{howard2017mobilenets} &	Triplet loss  \cite{wang2016deep,li2015feature} &	Hamming K-NN /Database &	Nordland  \cite{olid2018single}; \newline Gardens Point  \cite{sunderhauf2015performance} \\
				& \cite{8967783} &	 HybridNet \cite{chen2017deep} 	& Triplet Loss & Cosine /Database	& Oxford RobotCar \cite{barnes2020oxford}; Nordland \cite{olid2018single};  Gardens Point \cite{sunderhauf2015performance}; \\ 
				\midrule
				\multirow{8}{*}{\rotatebox[origin=c]{90}{3D LiDAR}} & \cite{liu2019lpd}	& \textbf{LPD-Net:}  \newline{} Adaptive feature extraction + a graph-based neighborhood aggregation + NetVLAD layer & 	Lazy quadruplet loss	& L2 /Database &	 Oxford Robotcar \cite{RobotCarDatasetIJRR};\\
				& \cite{angelina2018pointnetvlad} &\textbf{PointNetVLAD:}  \newline{} PointNet + NetVLAD layer &	Lazy triplet and quadruplet loss & KNN /Database &	Oxford RobotCar \cite{RobotCarDatasetIJRR};\\
				& \cite{8968094} & \textbf{OREOS:} \newline{} CNN as in \cite{simonyan2014very,appalaraju2017image} &	 Triplet loss \cite{hoffer2015deep}	& KNN /Database	& NCLT \cite{carlevaris2016university}; \newline{}  KITTI   \cite{Geiger2012CVPR};\\
				& \cite{8500682} & \textbf{LocNet:} \newline{} Siamese network & Contrastive loss function\cite{1640964}	& L2 KNN /Database & KITTI  \cite{Geiger2012CVPR} \newline{} inhouse dataset\\ 
				& \cite{8734150} &	Siamese network	 & Contrastive loss \cite{1640964}	& L2  KNN /Database	& KITTI  \cite{Geiger2012CVPR}; inhouse dataset;\\
				\midrule
				RADAR &	\cite{suaftescu2020kidnapped}	&  VGG-16 + NetVLAD layer &	Triplet loss &	KNN /Database &	Oxford Radar RobotCar \cite{barnes2020oxford} \\
				\bottomrule
			\end{tabular} 
		\end{adjustbox}
	}
	\label{tab:end-to-end}%
\end{table*}%

\subsection{End-to-End Frameworks}
\label{sec:endtoend}

Conversely to pre-trained frameworks, end-to-end frameworks resort to machine learning approaches that learn the feature representation and obtain a descriptor directly from the sensor data while training on a place recognition task.
A key aspect of end-to-end learning is concerned with the definition of the training objective: i.e., what are the networks optimized for, and how are they optimized. In place recognition, networks are mostly optimized to generate unique descriptors that can identify the same physical place regardless of the appearance or viewpoint. The achievement of such an objective is determined by selecting, for the task in hands, an adequate network , and adequate network training, which depends on the loss function.

\subsubsection{Loss functions}

The loss function is in particular a major concern in the training phase, since it represents the matematical interpretation of the training objective, and thus determining the successful convergence of the optimization process. In place recognition loss  functions include triplet-based \cite{arandjelovic2016netvlad,yu2019spatial,8967783,angelina2018pointnetvlad,8968094,suaftescu2020kidnapped,8968599},  margin-based\cite{9140362},  quadruplet-based \cite{liu2019lpd}, and  contrastive-based \cite{8500682}. Figure \ref{fig:loss} illustrates the various training strategies of the loss functions. 

The contrastive loss  is used in siamese networks \cite{8500682,1640964}, which have two branches with shared parameters. This function computes the similarity distance between the output descriptors of the branches, forcing the netwroks to decrease the distance between positive pairs (input data from the same place) and increase the distance between negative pairs. The function can be described as follow:

\begin{equation} \label{eq:contrastive_loss}
    \begin{gathered}
        L =  \frac{1}{2}  Y D^2  +\frac{1}{2}  (1-Y)\max(0,m-D)^2  
    \end{gathered}
\end{equation}

\noindent where $D = || R_a - R_x||_2$ represents the Euclidean distance between the descriptor representation from the branch of the anchor image ($R_a$)  and the descriptor representation from the other branch ($R_x$). While $m$ represents a margin parameter, $Y$ represents the label, where $Y=1$ refers to a positive pair and $Y=0$ otherwise. 

Similar to the former loss, the triplet loss also relies on more than one branch during training. However, instead of computing the distance between positive or negative pairs at each iteration, the triplet loss function computes the distance between a positive and a negative pair at the same iteration, relying, thus, on three branches.  As in the former loss function, the objective is to train a network to keep positive pairs close and negative pairs apart. The Triplet loss function can be formulated as follows: 

\begin{equation} \label{eq:triple_loss}
        L= \max(0,D_p^2 - D_n^2 + m)
 \end{equation}

\noindent where $D_p$ refers to the distance between positive pairs (i.e., between anchor and positive sample) and $D_n$ refers to the distance between the negative pair. This function is widely used in place recognition, namely in frameworks that use input data from the camera, 3d LiDARs, and RADARs, which adapt the function to fit the training requirements. Loss functions that drive from the triplet loss include Lazy triplet  \cite{angelina2018pointnetvlad}, weighted triplet loss \cite{yu2019spatial} and weakly supervised triplet ranking loss \cite{arandjelovic2016netvlad}.  

The quadruplet is an extension of the triplet loss, introducing an additional constraint to push the negative pairs \cite{chen2017beyond} from the positives pairs w.r.t different probe samples, while triplet loss only pushes the negatives from the positives w.r.t from the same probe. The additional constrain of the quadruplet loss reduces the intra-class variations and enlarges the inter-class variations. This function is formulated as follows:
\begin{equation} \label{eq:quadroplet_loss}
    \begin{gathered}
        L = \max(0,D_p^2 - D_n^2 + m_1) + \max(0,D_p^2 - D_d^2 + m_2)
    \end{gathered}
\end{equation}
 
\noindent where $D_p$ and $D_n$ represent the distance between the positive and negative pairs, respectively. The $m_1$ and $m_2$ represent margin parameters, while $R_d$ corresponds to the additional constraint, representing the distance between negative pairs from different probes.  In \cite{liu2019lpd,angelina2018pointnetvlad}, the quadruplet loss function is used to train networks for the task of place recognition using 3D LiDAR data.

The margin-based loss function is a simple extension to the contrastive loss \cite{8237571}.  While the contrastive function enforces the positive pairs to be as close as possible, the margin-based function only encourages the positive pairs to be within a distance of each other.  
    \begin{equation}
        L = max(0,\alpha + Y (D - m))
    \end{equation}

\noindent where $Y \in {1,-1}$ represents the labels of $Y=1$ when the pair is positive and $Y = -1$ otherwise.  $\alpha$ is a variable that determines the boundary between positive and negative pairs.  The margin-based loss function was proposed in \cite{8237571} to demonstrate that state-of-art performance could be achieved with a simple loss function, only by having an adequate sampling strategy of the input data during training.    This function is used in \cite{9140362}  to train a multi-modal network. The network is jointly trained based on information extracted from images and structural data in the format of voxel grids, which are generated from the images.

\subsubsection{Camera-based Networks}

A key contribution to supervised end-to-end-based place recognition is the NetVLAD layer \cite{arandjelovic2016netvlad}. Inspired by the Vector of Locally Aggregated Descriptors (VLAD) \cite{jegou2010aggregating}, Arandjelovi\'{c} et al. \cite{arandjelovic2016netvlad}  propose NetVLAD as a `pluggable' layer into any CNN architecture to output a  compact image descriptor. The network's parameters are learned using a weakly supervised triplet ranking loss function. 

Yu et al. \cite{yu2019spatial} also exploited VLAD descriptors for images, proposing a spatial pyramid-enhanced VLAD (SPE-VLAD) layer. The proposed layer leverages the spatial pyramid structure of images to enhance place description, using for feature extraction a VGG-16 \cite{simonyan2014very} and a ResNet-18 \cite{he2016deep},  and as loss function the weighted T-loss. The network's parameters are learned under weakly supervised scenarios, using GPS tags and the Euclidean distance between the image representations.

Qiu et al. \cite{qiu2018siamese} apply a siamese-based network to loop closure detection. Siamese networks are twin neural networks, which share the same parameters and are particularly useful when limited training data is available. Both sub-networks share the same parameters and mirror the update of the parameters. In this work, the sub-networks are replaced by ResNet to improve feature representation, and the network is trained, resorting to an L2-based loss function as in \cite{chopra2005learning}.

Wu et al. \cite{8968599} jointly addresses the place recognition problem from the efficiency and performance perspective, proposing to this end a deep supervised hashing approach with a similar hierarchy. Hashing is an encoding technique that maps high dimensional data into a set of binary codes, having low computational requirements and high storage efficiency. The proposed framework comprises three modules: features extraction based on MobileNet \cite{howard2017mobilenets}; hash code learning, obtained using the last fully connected layer of MobileNet; and loss function, which is based on the likelihood \cite{wang2016deep,li2015feature}. This work proposes a similar hierarchy method to distinguish similar images. To this end, the distance of hashing codes between a pair of images must increase as similar images are more distinct and must remain the same between different images. These two conditions are essential to use deep supervised hashing in place recognition. 

Another efficiency improving technique for deep networks is network pruning. This technique aims to reduce the size of the network by removing unnecessary neurons or setting the weights to zero \cite{blalock2020state}.  Hausler et al. \cite{8967783} propose a feature filtering approach, which removes feature maps at the beginning of the network while using for matching late feature maps to foster efficiency and performance simultaneously.  Feature maps to be removed are determined based on a Triplet Loss calibration procedure. As a feature extraction framework, the approach uses the HybridNet \cite{chen2017deep}.

Contrary to former single modality works, Oertel et al. \cite{9140362} propose a place description approach that uses vision and structural information, both originated from camera data. This approach jointly uses vision and depth data from a stereo camera in an end-to-end pipeline. The structural information is first obtained utilizing the Direct Sparse Odometry (DSO) framework \cite{engel2017direct} and then discretized into regular voxel grids to serve as inputs along with the corresponding image. The pipeline has two parallel branches: one for vision and another for the structural data, which use 2D and 3D convolutional layers, respectively, for feature extraction. Both branches are learned jointly through a margin-based loss function. The outputs of the branches are concatenated into a single vector, which is fed to a fusion network that outputs the descriptor.

\subsubsection{3D LiDAR-based Network}

Although NetVLAD was originally used for images, it has also been used on 3D LiDAR data \cite{angelina2018pointnetvlad,liu2019lpd}.  Uy et al. \cite{angelina2018pointnetvlad} and Liu et al. \cite{liu2019lpd}  propose respectively PointNetVLAD and LPD-Net, which are NetVLAD-based global descriptor learning approaches for 3D LiDAR data. Both have compatible inputs and outputs, receiving as input raw point clouds and outputting a descriptor.  The difference relies on the feature extraction and feature processing methods. PointNetVLAD \cite{angelina2018pointnetvlad} relies on PointNet \cite{qi2017pointnet}, a  3D object detection and segmentation approach, for feature extraction. In contrast, LPD-Net relies on an adaptive local feature extraction module and a graph-based neighborhood aggregation module, aggregating both in the Feature Space and the Cartesian Space. Regarding the network training, Uy et al. \cite{angelina2018pointnetvlad} showed that the lazy quadruplet loss function enables higher performance than the lazy triplet loss function, motivating Liu et al. \cite{liu2019lpd} to follow this approach.

A different 3D-LiDAR-based place recognition approach is proposed in \cite{8968094}. Schaupp et al. propose OREOS, which is a triplet DL network-based architecture \cite{hoffer2015deep}. The OREOS approach receives as input 2D range images and outputs orientation-and place-dependent descriptors. The 2D range images are the result of the 3D point clouds projections onto an image representation. The network is trained using an L2 distance-based triplet loss function to compute the similarity between anchor-positive and anchor-negative. Place recognition is validated using a k-nearest neighbor framework for matching.

Yin et al.	\cite{8500682} uses  3D point clouds to address the global localization problem, proposing a place recognition and metric pose estimation approach.  Place recognition is achieved using the siamese LocNets, which is a semi-handcrafted representation learning method for LiDAR point clouds.  As input, LocNets receives a handcrafted rotational invariant representation extracted from point clouds in a pre-processing step and outputs a low-dimensional fingerprint. The network follows a Siamese architecture and uses for learning Euclidean distance-based contrastive loss function \cite{1640964}. For belief generation, an L2-based  KNN approach is used. A similar LocNets-based approach is proposed in \cite{8734150}.

\subsubsection{RADAR-based}

Regarding RADAR-based place recognition, Saftescu et al. \cite{DBLP:conf/icra/SaftescuGMBN20} also propose a NetVLAD-based approach to map FMCW RADAR scans to a descriptor space. Features are extracted using a specially tailored CNN  based on cylindrical convolutions, anti-aliasing blurring, and azimuth-wise max-pooling to bolster the rotational invariance of polar radar images.  Regarding training,  the network uses a triplet loss function as proposed in \cite{schroff2015facenet}.

\section{Unsupervised Place Recognition} \label{sec:unsupervised}

The aforementioned supervised learning approaches achieve excellent results in learning discriminative place models. However, these methods have the inconvenience of requiring a vast amount of labeled data to perform well, as it is common in supervised DL-based approaches. Contrary to supervised, unsupervised learning does not require labeled data, an advantage when annotated data are not available or scarce. 

Place recognition works use unsupervised approaches such as  Generative Adversarial Networks  (GAN) for domain translation \cite{latif2018addressing}. An example of such an approach is proposed by Latif et al. \cite{latif2018addressing}, which address the cross-season place recognition problem as a domain translation task. GANS are used to learn the relationship between two domains without requiring cross-domain image correspondences.  The proposed architecture is presented as two coupled GANs. The generator integrated an encoder-decoder network, while the discriminator integrates an encoder network followed by two fully connected layers. The output of the discriminator is used as a descriptor for place recognition. Authors show that the discriminator's feature space is more informative than image pixels translated to the target domain.   

Yin et al. \cite{8593562} also proposes a GAN-based approach, but for  3D LiDAR-based. LiDAR data are first mapped into dynamic octree maps, from which bird-view images are extracted. These images are used in a GAN-based pipeline to learn stable and generalized place features. The network trained using adversarial and conditional entropy strategies to produce a higher generalization ability and capture the unique mapping between the original data space and the compressed latent code space. 

Han et al.,(2020) \cite{9197514} propose a Multispectral Domain Invariant framework for the translation between unpaired RGB and thermal imagery. The proposed approach is based on CycleGAN \cite{8237506}, which relies, for training, on the single scale structural similarity index (SSIM \cite{1284395}) loss,  triplet loss, adversarial loss, and two types of consistency losses (cyclic loss \cite{8237506} and pixel-wise loss). The proposed framework is further validated on semantic segmentation and domain adaptation tasks.  

Contrary to the former works, which were mainly based on GAN approaches, Merril and Huang \cite{merrill2018lightweight} propose, for visual loop closure, an autoencoder-based approach to handle the feature embedding.  Instead of reconstructing original images, this unsupervised approach is specifically tailored to map images to a HOG descriptor space. The autoencoder network is trained, having as input a pre-processing stage, where two classical geometric vision techniques are exploited: histogram of oriented gradients (HOG) \cite{1467360}, and the projective transformation (homography) \cite{hartley2003multiple}.  HOG enables the compression of images while preserving salient features. On the other hand, the projective transformation allows the relation of images with differing viewpoints. The network has a minimal architecture, enabling fast and reliable close-loop detection in real-time with no dimensionality reduction.

\begin{table*}[t]
  \centering
  \caption{Summary of recent works using  unsupervised end-to-end learning techniques for place recognition.}
    \begin{adjustbox}{max width=\textwidth}
    {\renewcommand{\arraystretch}{2}
    \begin{tabular}{p{0.06\textwidth}p{0.04\textwidth}p{0.3\textwidth}p{0.2\textwidth}p{0.2\textwidth}p{0.2\textwidth}}
    \toprule
    \multicolumn{1}{c}{\textbf{Sensor}} & \multicolumn{1}{c}{\textbf{Ref}} & \multicolumn{1}{c}{\textbf{Architecture}} & \multicolumn{1}{c}{\textbf{Loss Function}} & \multicolumn{1}{c}{\textbf{Task}} & \multicolumn{1}{c}{\textbf{Dataset}} \\
    \midrule
    \midrule
    Camera & \cite{latif2018addressing}	 &	Architecture: Coupled GANs + encoder-decoder network	& Minimization of the cyclic reconstruction loss \cite{kim2017learning}	& Domain translation for cross domain place recognition &	 Norland   \cite{olid2018single}; \\
    Camera & \cite{merrill2018lightweight}	& Pre-processing:  HOG \cite{1467360} and homography \cite{hartley2003multiple} \newline  Architecture: small Autoencoder &	L2 loss function &	Unsupervised feature embedding for visual loop closure &	Places \cite{zhou2017places}; KITTI  \cite{Geiger2012CVPR}; Alderley \cite{milford2012seqslam}; Norland  \cite{olid2018single}; Gardens Point \cite{sunderhauf2015performance};\\
    RGB + Thermal	 & \cite{9197514} &  Multispectral Domain Invariant model \newline Architecture: CycleGAN \cite{8237506} &	SSIM \cite{1284395}  + triplet + adversarial + cyclic loss \cite{8237506} +  pixel-wise loss & Unsupervised multispectral imagery translation task & KAIST \cite{7298706}; \\
    \midrule
    3D LiDAR & \cite{latif2018addressing}	& Pre-processing: Mapping LiDAR to  dynamic octree maps to  bird-view images \newline Architecture: GAN + encoder-decoder network & Adversarial learning and  conditional entropy &	Unsupervised Feature learning for a 3D LiDAR-based place recognition task & KITTI \cite{Geiger2012CVPR}; \newline NCTL  \cite{carlevaris2016university}; \\
    \bottomrule
    \end{tabular}%
    }
    \end{adjustbox}
  \label{tab:unsupervised}%
\end{table*}%

\begin{table*}[t]
  \centering
  \caption{Recent works that combine supervised and unsupervised learning in place recognition systems.}
    \begin{adjustbox}{max width=\textwidth}
    {\renewcommand{\arraystretch}{2}
    \begin{tabular}{p{0.06\textwidth}p{0.04\textwidth}p{0.3\textwidth}p{0.2\textwidth}p{0.2\textwidth}p{0.2\textwidth}}
    \toprule
    \multicolumn{1}{c}{\textbf{Sensor}} & \multicolumn{1}{c}{\textbf{Ref}} & \multicolumn{1}{c}{\textbf{Architecture}} &\multicolumn{1}{c}{ \textbf{Loss Function}} & \multicolumn{1}{c}{\textbf{Task}} & \multicolumn{1}{c}{\textbf{Dataset}} \\
    \midrule
    \midrule
    \multirow{4}{*}{\rotatebox[origin=c]{90}{Camera}} &	\cite{9022242} & Feature extraction: AlexNet cropped(conv5) \newline Supervised: VLAD  +  attention module \newline Unsupervised:  domain adaptation & Supervised: triplet ranking; \newline Unsupervised: MK-MMD \cite{borgwardt2006integrating} &	Single and cross-domain VPR  &  Mapillary \footnotemark 
    \newline Beeldbank \footnotemark
    \\
    & \cite{9196518} & Supervised: adversarial learning \newline Unsupervised: autoencoder & Adversarial Learning: Least square \cite{8237566};  \newline Reconstruction: L2 distance & Disentanglement of  place and appearance features in a cross domain VPR & Nordland \cite{olid2018single}; \newline Alderley \cite{milford2012seqslam}; \\
    \midrule
    3D LiDAR &	\cite{dube2020segmap} &	Supervised:  latent space + classification network  \newline Unsupervised: Autoencoder-like network  & Classification: softmax cross entropy  \cite{parkhi2015deep} \newline Reconstruction: binary cross entropy \cite{brock2016generative} &	Global localization, 3D dense map reconstruction, and semantic information extraction &	KITTI odometry \cite{Geiger2012CVPR}; \\
    \bottomrule
    \end{tabular}%
    }
    \end{adjustbox}
  \label{tab:semi}%
\end{table*}%

\footnotetext[1]{https://www.mapillary.com} 
\footnotetext[2]{https://beeldbank.amsterdam.nl/beeldbank}

\section{Semi-supervised Place Recognition} \label{sec:semisupervised}

In this work, Semi-supervised approaches refer to works that jointly rely on supervised and unsupervised methods. The combination of these two learning approaches is particularly used for the cross-domain problem. However, rather than translating one domain to another, these learning techniques are used to learn features that are independent of the domain appearance. A summary of recent works is presented in Table \ref{tab:semi}.    

To learn domain-invariant features for cross-domain visual place recognition, Wang et al. \cite{9022242} propose an approach that combines weakly supervised learning with unsupervised learning. The proposed architecture has three primary modules: an attention module,  an attention-aware VLAD module, and a domain adaptation module. The supervised branch is trained with a triplet ranking loss function, while the unsupervised branch resorts to a multi-kernel maximum mean discrepancy  (MK-MMD) loss function.

On the other hand, Tang et al. \cite{9196518} propose a self-supervised learning approach to disentangle place-rated features from domain-related features. The backbone architecture of the proposed approach is a modified autoencoder for adversarial learning: i.e., two input encoder branches converging into one output decoder. The disentanglement of the two feature domains is solved through adversarial learning, which constrains the learning of domain specific features (i.e., features depending on the appearance); a task that is not guaranteed by the reconstruction loss of autoencoders. For adversarial learning, the proposed loss function is the least square adversarial loss \cite{8237566}; while for reconstruction, the loss function is the L2 distance.

Dubé et al. \cite{dube2020segmap} propose SegMap, an data-driven learning approach for the task of localization and mapping. The approach uses as the main framework an autoencoder-like architecture to learn object segments of 3D point clouds. The framework is used for two tasks: (supervised) classification and (unsupervised) reconstruction. The work proposes a customized learning technique to train the network, which comprises, for classification, the softmax cross-entropy loss function in conjunction with the N-ways classification problem learning technique \cite{parkhi2015deep}, and, for reconstruction, the binary cross-entropy loss function \cite{brock2016generative}. The latent space, which is jointly learned on the two tasks, is used as a descriptor for segment retrieval. The proposed framework can be used in global localization, 3D dense map reconstruction, and semantic information extraction tasks.

\section{Other Frameworks} \label{sec:other}

 \begin{figure*}[t]
	\centering
	\includegraphics[width=1\linewidth]{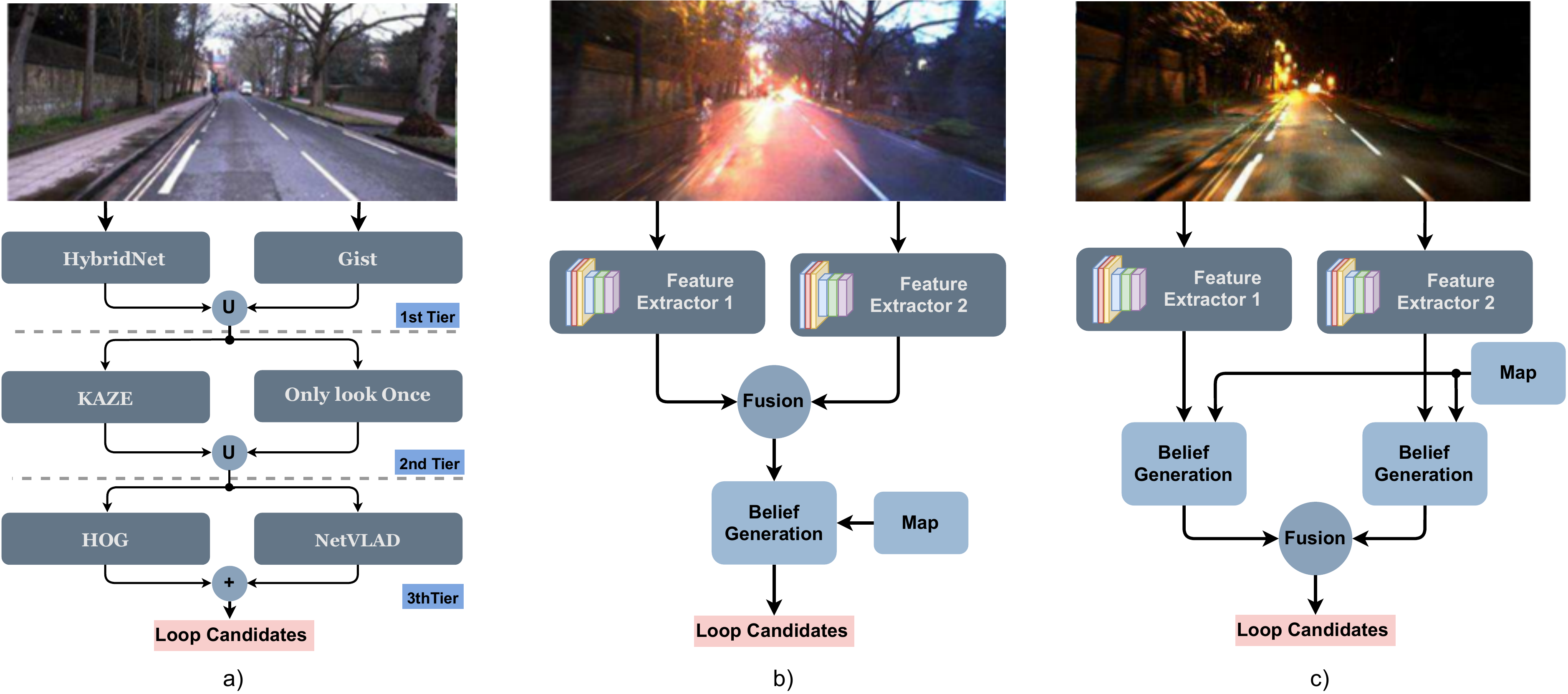}
	\caption{Block diagram of a) hierarchical b) and c) parallel place recognition frameworks. The example in b) fuses the descriptors, while in c) the belief scores are fused.}
	\label{fig:other}	
\end{figure*}

This section is dedicated to the frameworks that have more complex and entangled architectures: i.e., containing more than one place recognition approach for the purpose of finding the best loop candidates.  Two main frameworks are highlighted: parallel and hierarchical.  While parallel frameworks have a very defined structure, hierarchical frameworks may assume very complex and entangled configurations, but both frameworks have the end goal of representing more performant place recognition methods. 

\subsection {Parallel Frameworks}

Parallel frameworks refer to approaches that rely on multiple information streams, which are fused into one branch to generate place recognition decisions.  These parallel architectures fuse the various branches utilizing methods such as feature concatenation  \cite{zhang2016robust}, HMM \cite{hausler2019multi} or multiplying normalized data across Gaussian-distributed clusters \cite{jacobson2018leveraging}. Approaches such as proposed by Oertel et al. \cite{9140362}, where vision and structural data are fused in an end-to-end fashion, are considered to belong to the Section \ref{sec:endtoend} because features are jointly learned in an end-to-end pipeline. An example of parallel frameworks is illustrated in Fig. \ref{fig:other}, and a summary of recent works is presented in Table \ref{tab:parallel}.

Relying on multiple information streams allows overcoming individual sensory data limitations, which can be, for instance, due to environment changing conditions. Zhang et al. \cite{zhang2016robust} address the loop closure detection problem under strong perceptual aliasing and appearance variations, proposing Robust Multimodal Sequence-based (ROMS). ROMS concatenates LDB features \cite{Arroyo2015towards}, GIST features\cite{Latif2014an}, CNN-based deep features\cite{ren2015faster} and ORB local features \cite{mur2015orb} in a single vector.  A similar (parallel) architecture is proposed by Hausler et al. \cite{hausler2019multi}, where an approach, called Multi-Process Fusion, fuses four image processing methods: SAD with patch normalization \cite{milford2012seqslam,pepperell2014all}; HOG \cite{naseer2018robust,McManus-RSS-14}; multiple spatial regions of CNN features \cite{chen2017deep,chen2018learning}; and spatial coordinates of maximum CNN activations \cite{garg2018lost}. However, instead of fusing all features to generate one descriptor as proposed in \cite{zhang2016robust}, here, each feature stream is matched separately using cosine distance, and only the resulting similarity values are fused using the Hidden Markov model.

\begin{table*}[t]
	\centering
	\caption{Summary of recent works on supervised place recognition using parallel frameworks.  BG = Belief Generation and PM = Place mapping. }
	{\renewcommand{\arraystretch}{2}
		\begin{adjustbox}{max width=\textwidth}
			\begin{tabular}{p{0.06\textwidth}p{0.04\textwidth}p{0.45\textwidth}p{0.2\textwidth}p{0.15\textwidth}p{0.2\textwidth}}
				\toprule
				\multicolumn{1}{c}{\textbf{Sensor}} & \multicolumn{1}{c}{\textbf{Ref}} & \multicolumn{1}{c}{\textbf{Model}} &\multicolumn{1}{c}{\textbf{Fusion}} & \multicolumn{1}{c}{\textbf{BG/PM}} & \multicolumn{1}{c}{\textbf{Dataset}}\\
				\midrule
				\midrule
				\multirow{4}{*}{\rotatebox[origin=c]{90}{Camera}} & \cite{zhang2016robust} &	Feacture Extraction: LDB \cite{Arroyo2015towards}  + GIST \cite{Latif2014an} + CNN \cite{ren2015faster} + ORB \cite{mur2015orb} &  Concatenation of all features &	Sequence  /Database	& St Lucia  \cite{glover2010fab};  CMU-VL \cite{badino2012real};  Nordland \cite{olid2018single}; \\
				&	\cite{hausler2019multi} & Feacture Extraction: SAD \cite{milford2012seqslam,pepperell2014all} + HOG \cite{naseer2018robust,McManus-RSS-14} +  spatial regions of  HybridNet(Conv-5 layer)  \cite{chen2017deep,chen2018learning} +  spatial coordinates of maximum activations HybridNet(Conv-5 layer) \cite{garg2018lost} \newline Descriptor: features + normalization & Hidden Markov Model of the similarity distances of each feature stream	  & Dynamic sequence /Database & St Lucia \cite{glover2010fab} \newline Nordland \cite{olid2018single} \newline Oxford RobotCar  \cite{philbin2007object}\\				
				\bottomrule
			\end{tabular} 
		\end{adjustbox}
	}
	\label{tab:parallel}%
\end{table*}%

\subsection {Hierarchical Frameworks}

In this work, hierarchical frameworks refer to place recognition approaches that, similarly to parallel frameworks, rely on multiple methods; however, instead of having as main framework a parallel architecture, the architecture is formed by various stacked tiers. The hierarchical architectures find the best loop candidate by filter candidates progressively in each tier. An example of such a framework is the coarse-to-fine architecture, which has a coarse and a fine tier. The coarse tier is mostly dedicated to retrieving top candidates utilizing methods that rather are computer efficient than accurate. These top candidates are feed to the fine tier, which can use more computer demanding methods to find the best loop candidate. The coarse-to-fine architecture, while being the most common, is not the only. Other architectures exist, for example Fig. \ref{fig:other} illustrates a framework proposed in \cite{DBLP:conf/icra/HauslerM20} and Table \ref{tab:hierarchical} presents a summary of recent works. 

Hausler and Milford \cite{DBLP:conf/icra/HauslerM20} show that parallel fusion strategies have inferior performance compared with hierarchical approaches and therefore propose Hierarchical Multi-Process Fusion, which has a three-tier hierarchy. In the first tier, top candidates are retrieved from the database based on HybridNet\cite{chen2017deep} and   Gist\cite{oliva2001modeling} features. In the second tier,  from the top candidates of the previous tier, a more narrow selection is performed based on  KAZE\cite{alcantarilla2012kaze} and  Only Look Once (OLO)\cite{chen2017only} features. Finally, the best loop candidate is obtained in the third tier using NetVLAD\cite{arandjelovic2016netvlad} and HOG\cite{dalal2005histograms}. An illustration of this framework is presented in Fig. \ref{fig:other}.

Garg et al. \cite{garg2018lost} also follow a similar framework, proposing a hierarchical place recognition approach, called X-Lost. In the coarse tier, top candidates are found by matching the Local Semantic Tensor (LoST) descriptor, which comprises feature maps from the  RefineNet \cite{lin2017refinenet} (a dense segmentation network) and semantic label scores of the road, building, and vegetation classes. The best match is found in the fine tier by verifying the spatial layout of semantically salient keypoint correspondences.    

This semantic- and keypoint-based approach is further exploited in  \cite{garg2019semantic,garg2019look}. In \cite{garg2019semantic}, top candidates are obtained fusing NetVLAD\cite{arandjelovic2016netvlad} and LoST\cite{garg2018lost} descriptors in a coarse stage, while in  \cite{garg2019look}, depth maps are computed  from camera data in an intermediate stage to remove keypoints that are out of range. 

Contrary to the former approaches, where performance is the primary goal, An et al.\cite{8968043} address the efficiency problem of place recognition, proposing an approach based on an HNSW graph for efficient map management.   HNSW graph guarantees low map building and retrieval time. In the coarse stage, top candidates are retrieved from the HNSW graph by matching features extracted from the MobileNetV2 \cite{sandler2018mob} using normalized scalar product \cite{sivic2008efficient}. The final loop candidate is obtained by matching hash codes from SURF features and the top candidates retrieved in the coarse stage.   
On the other hand, Liu et al. \cite{liu2019seqlpd}  exploits 3D point clouds instead of camera data, proposing SeqLPD, which is a lightweight variant of our LPD-Net \cite{liu2019lpd}. This approach resorts to super keyframe clusters for coarse search, while for fine search, local sequence matching is preferred.

\begin{table*}[t]
	\centering
	\caption{Summary of recent works using hierarchical frameworks.  BG = Belief Generation and PM = Place mapping. }
	{\renewcommand{\arraystretch}{2}
		\begin{adjustbox}{max width=\textwidth}
			\begin{tabular}{p{0.05\textwidth}p{0.04\textwidth}p{0.4\textwidth}p{0.4\textwidth}p{0.15\textwidth}p{0.2\textwidth}}
				\toprule
				\multicolumn{1}{c}{\textbf{Sensor}} & \multicolumn{1}{c}{\textbf{Ref}} & 
				\multicolumn{1}{c}{\textbf{Coarse Stage}} &
				\multicolumn{1}{c}{\textbf{Fine Stage}} & \multicolumn{1}{c}{\textbf{PM}} & \multicolumn{1}{c}{\textbf{Dataset}} \\
				\midrule
				\midrule
				\multirow{15}{*}{\rotatebox[origin=c]{90}{Camera}} & \cite{garg2018lost} & Features:  RefineNet \cite{lin2017refinenet} \newline Descriptor: LosT (semantic label scores +  conv5 layer feature maps ) + normalization \newline  BG: cosine distance & Features:  RefineNet \cite{lin2017refinenet} \newline Descriptor:  keypoint  extracted from CNN layer activations \newline  GB: spatial layout Verification ( Semantic Label Consistency + weighted Euclidean distance) & Coarse: Database \newline Fine: Top Candidates  & Oxford Robotcar \cite{RobotCarDatasetIJRR}; \newline  Synthia Dataset \cite{7780721}; \\
				& \cite{garg2019semantic} & Descriptor: concatenation of LoST \cite{garg2018lost} + NetVLAD\cite{arandjelovic2016netvlad} \newline BG:  cosine distance &	Features:  pre-trained CNN \newline  Descriptor:  keypoint  extracted from CNN activations \newline  BG: spatial layout consistency & Coarse: Database \newline Fine: Top Candidates &	Oxford Robotcar \cite{RobotCarDatasetIJRR}; MLFR;  Parking Lot; Residence Indoor Outdoor \\
				& \cite{garg2019look}& Features: RefineNet(Resnet101) \cite{lin2017refinenet}\newline Descriptor: conv 5 feature maps \newline BG:  cosine distance &	Filtering: out-of-range  keypoins  based on depth maps \newline Descriptor: same as coarse stage (filtered) \newline BG: cosine distance &  Coarse: Database \newline Fine: Top Candidates & Oxford Robotcar \cite{RobotCarDatasetIJRR} \newline Synthia  \cite{7780721} (for depth evaluation) \\
				& \cite{8968043} &	Top Candidates:\newline  Features: MobileNetV2  \cite{sandler2018mob} \newline  Descriptor:  final average pooling layer \newline BG: nearest neighbors + normalized scalar product \cite{sivic2008efficient} & Features:  SURF  \newline Descriptor: hash codes  \newline BG: Hamming Distance between top candidates and SURF-based descriptor + ratio test \cite{lowe2004distinctive}  + RANSAC & Coarse: HNSW graphs \newline  Fine: Top candidates & KITTI \cite{Geiger2012CVPR} \newline  Malaga 2009 Parking 6L \cite{blanco2009collection}  \newline  New College \cite{smith2009new} \\
				& \cite{DBLP:conf/icra/HauslerM20} &  1st Tier: \newline Features: HybridNet(AlexNet) \cite{chen2017deep} and Gist \cite{oliva2001modeling} \newline BG: Difference Scores + normalization \newline 2nd Tier:\newline Features: KAZE  \cite{alcantarilla2012kaze} and Only Look Once \cite{8202131} \newline BG: (KAZE) sum of the residual distances and difference scores + normalization  & 3 Tier: \newline Features: NetVLAD  \cite{arandjelovic2016netvlad} and HOG \cite{dalal2005histograms} \newline BG: max(Average of  Difference Scores) & 1st tier: Database \newline 2nd tier: Top candidates of the 1st tier \newline 3th tier: Top candidates of the 2nd tier &  Nordland \cite{niko2013we} \newline Berlin Kurfurstendamm \cite{sunderhauf2015place} \\
				\midrule
				3D LiDAR & \cite{liu2019seqlpd} & Find the cluster: \newline Descriptor: lightweight variant LPD-Net  \newline  Matching: the nearest L2 distance to the cluster center is selected as the super keyframe &	 Descriptor: same as  in coarse \newline BG: Local sequence matching	& Super keyframe  clusters & Oxford Robotcar \cite{RobotCarDatasetIJRR}  \newline KITTI   \cite{Geiger2012CVPR} \\
				\bottomrule
			\end{tabular} 
		\end{adjustbox}
	}
	\label{tab:hierarchical}%
\end{table*}%

\section{Conclusion and discussion}
\label{sec:conclusion}

This paper presents a critical survey on place recognition approaches, emphasizing the recent developments on deep learning frameworks, namely supervised, unsupervised, semi-supervised, parallel, and hierarchical approaches.

An overview of each of these frameworks is presented. In supervised approaches, the pre-trained frameworks tend to resort to semantic information by detecting landmarks or leveraging regional activation from CNN layers. On the other hand, among the end-to-end frameworks, the NetVLAD layer has inspired various works, which integrated this layer in deep architectures to train the model directly on place recognition using sensory data from the camera, 3D LiDAR, or RADAR. The main application of unsupervised approaches, such as GANs and autoencoders, is to address the domain translation problem.  While in semi-supervised, which in this work refers to works that jointly leverage supervised and unsupervised methods, the works address the cross-domain problem, however instead of translating a source domain into a target domain, these works seek to obtain a descriptor space that is invariant to domains.  Besides these traditional machine learning frameworks, other frameworks have been suggested, combining multiple DL or classical ML approaches into a parallel or hierarchical architecture.  In particular, the hierarchical approach has been shown to improve performances in general. Until recently, the primary motivation of the majority of the published articles was to increase performance. However, recent works additionally to high performance are also seeking efficiency.

\section*{Acknowledgments}
This work has been supported by the projects MATIS-CENTRO-01-0145-FEDER-000014 and SafeForest CENTRO-01-0247-FEDER-045931, Portugal. It was also partially supported by FCT through grant UID/EEA/00048/2019.

\bibliographystyle{IEEEtran}

\bibliography{references}


\setlength\intextsep{1pt}


\hfill 

\begin{wrapfigure}{l}{0.14\textwidth}
	\includegraphics[width=0.14\textwidth,trim={0cm 0cm 0cm 0cm}]{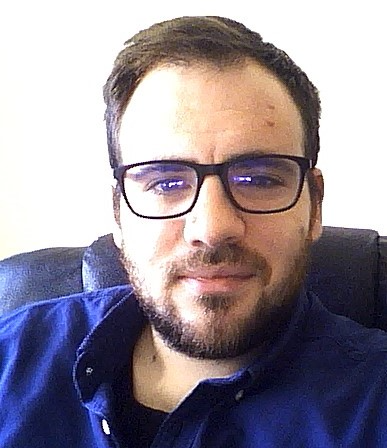}
\end{wrapfigure}

\noindent \textbf{Tiago Barros}  received the Master degree in Electrical and Computer Engineering from the University of Coimbra, Portugal, in 2015. He is currently working towards the Ph.D. degree in the Institute of Systems and Robotics, University of Coimbra, Portugal. His research interests include deep learning, perception and localization.  \\

\hfill \break 

\begin{wrapfigure}{l}{0.14\textwidth}
	\includegraphics[width=0.14\textwidth, trim={0.0cm 3cm 0cm 2.5cm},clip]{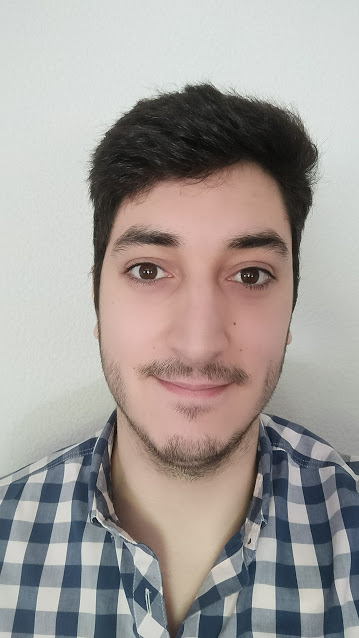}
\end{wrapfigure}

\noindent  \textbf{Ricardo Pereira} received the Master degree in Electrical and Computer Engineering from the University of Coimbra, Portugal. He is currently working towards the Ph.D. degree in the Institute of Systems and Robotics, University of Coimbra, Portugal. His research interests include deep learning, perception, and mobile robotics.\\


\begin{wrapfigure}{l}{0.14\textwidth}
	\includegraphics[width=0.14\textwidth]{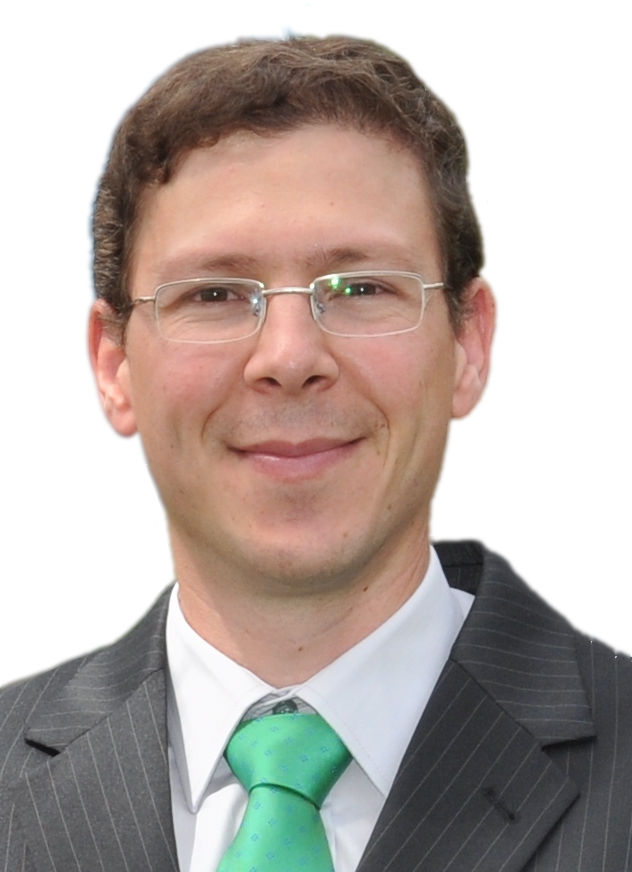}
\end{wrapfigure}

\noindent  \textbf{Cristiano Premebida} is Assistant Professor in the department of electrical and computer engineering at the University of Coimbra, Portugal, where he is a member of the Institute of Systems and Robotics (ISR-UC). His main research interests are autonomous vehicles, autonomous robots, robotic perception, cooperative/connected intelligent transport systems (CITS), ADAS, machine learning, and sensor fusion. 

C. Premebida has collaborated on research projects in the areas related to C-ITS, autonomous driving, and applied machine learning, including national and international projects. He is an IEEE-ITS society member, has served as AE in the flagship conferences ITSC and IVS, and has regularly organized international workshops on automated driving, AI/ML perception, and C-ITS. \\


\begin{wrapfigure}{l}{0.14\textwidth}
	\includegraphics[width=0.14\textwidth]{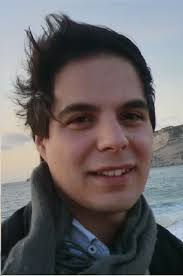}
\end{wrapfigure}

\noindent  \textbf{Lu\'{i}s Garrote} received the Master degree in Electrical and Computer Engineering from the University of Coimbra, Portugal. He is currently working towards the Ph.D. degree in the Institute of Systems and Robotics, University of Coimbra, Portugal. His research interests include deep learning, perception, and mobile robotics.\\


\begin{wrapfigure}{l}{0.14\textwidth}
	\includegraphics[width=0.14\textwidth]{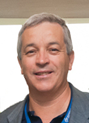}
\end{wrapfigure}

\noindent \textbf{Urbano J. Nunes} (S’90-M’95-SM’09) received the Ph.D. in Electrical Engineering from the University of Coimbra, Portugal, in 1995. He is a Full Professor with the Electrical and Computer Engineering Department of Coimbra University, and a Senior researcher of the Institute for Systems and Robotics (ISR-UC) where he is the coordinator of the Human-Centered Mobile Robotics lab. He has been involved with/responsible for several funded projects at both national and international levels in the areas of mobile robotics. He serves as Associate Editor the IEEE Transactions on Intelligent Vehicles (2015-). Prof. Nunes was with several international conferences: General co-chair of the 11th IEEE ICAR2003; Program Chair of IEEE ITSC2006; General Chair of the 13th IEEE ITSC2010; General Chair for the IEEE/RSJ IROS 2012; and General Chair of the IEEE ROMAN2017.

\end{document}